\documentclass{article}
\usepackage{spconf,amsmath,graphicx,amsfonts, bm, booktabs,amssymb}
\usepackage[ruled,linesnumbered]{algorithm2e}

\title{Efficient Client Contribution Evaluation \\for Horizontal Federated Learning}
\name{Jie Zhao\sthanks{Work done as an intern at Ping An Technology (Shenzhen) Co., Ltd. }$^{ \dagger}$ \qquad Xinghua Zhu$^{\star}$\sthanks{These authors contributed equally to this work.} \qquad Jianzong Wang$^{\star}$\sthanks{Corresponding author: jzwang@188.com.}\qquad Jing Xiao$^{\star}$}

\address{$^{\star}$ Ping An Technology (Shenzhen) Co., Ltd. \qquad $^{*}$ Hainan University}
    
\begin{document}
%
\maketitle
\begin{abstract}

In federated learning (FL), fair and accurate measurement of the contribution of each federated participant is of great significance. 
The level of contribution not only provides a rational metric for distributing financial benefits among federated participants,
but also helps to discover malicious participants that try to poison the FL framework.
Previous methods for contribution measurement were based on enumeration over possible combination of federated participants.
Their computation costs increase drastically with the number of participants or feature dimensions, making them inapplicable in practical situations.
In this paper an efficient method is proposed to evaluate the contributions of federated participants.
This paper focuses on the horizontal FL framework, where client servers calculate parameter gradients over their local data, and upload the gradients to the central server.
Before aggregating the client gradients, the central server train a data value estimator of the gradients using reinforcement learning techniques.
As shown by experimental results, the proposed method consistently outperforms the conventional leave-one-out method in terms of valuation authenticity as well as time complexity.
\end{abstract}
\section{Introduction}
\label{sec:intro}
In recent years, federated learning (FL) \cite{konevcny2016federated, kairouz2019advances, li2020knowledge} has received increasing attention in the machine learning society. 
In various machine learning tasks, FL allows the use of isolated data from multiple resources without violating the privacy protection policy \cite{zhu2019federated, kong2020network, zhu2020empirical}. 
At their basis, FL systems rely on the participation of individual data holders \cite{he2020fedsmart,konevcny2015federated}.
Computer scientists as well as economists are working closely to promote motivations for data holders to join in broader FL applications \cite{liu2020fedcoin,zhan2020learning}.
On one hand, economists provide pricing strategies for data contributors based on their gain and loss in the FL ecosystems.
On the other hand, computer scientists strive to achieve effective evaluation of the contributions from each data source \cite{wang2019measure}.
Therefore, a fair and accurate evaluation of the contributions of federated participants is important for completing a FL ecosystem \cite{zhang2020hierarchically}. 

The quality and quantity of data is centric to all machine learning algorithms. 
It has been shown that the performance of a neural network scales sublinearly with the size of the training set \cite{hestness2017deep}.
Series of methods have been proposed in the literature to determine the value of each data point in the training set.
Delete-and-retrain methods, such as leave-one-out (LOO) \cite{Koh2017UnderstandingBP} and Data Shapley \cite{ghorbani2019data} provided brute-force solutions to the valuation problem.
To ameliorate the computational expense, approximated LOO \cite{Koh2017UnderstandingBP} was proposed for faster but more constrained value estimation.
Data valuation using Reinforcement Learning (DVRL) \cite{yoon2019data}, which formulated data valuation as a meta learning framework, was recently proposed by Yoon et al.
In their experiments, DVRL outperformed LOO and Data Shapley algorithms in spotting corrupted data points.

In FL systems, 
Wang et al \cite{wang2019measure} adopted the deletion diagnostic methods, LOO and Data Shapley, to measure the participant contributions.
As the number of participants or the number of features increased, the computational cost increased drastically in Wang et al.'s methods.
These methods are hardly applicable in any practical configurations.

In this paper, an integrated client contribution evaluation method for the horizontal FL systems is proposed.
The proposed method is based on reinforcement learning (RL) and named Federated REINFORCE Client Contribution Evaluation (F-RCCE).
The reinforcement signal in F-RCCE is based on the performance of a small privately-held validation dataset of central server. 
F-RCCE can fairly and cost-effectively measure the contribution to a federated model by each client, in a privacy-preserving manner.
The main contributions of this study are:
\begin{enumerate}
    \vspace{-0.2cm}
    \item Propose a novel participant contribution evaluation method for FL systems, that uses RL to evaluate contributions fairly; 
    \vspace{-0.2cm}
    \item Verify the effectiveness and the time cost of the proposed method on the horizontal FL scenario;
    \vspace{-0.2cm}
    \item Investigation of the performance of FL systems under imbalanced data distributions using the proposed method.
\end{enumerate}

\vspace{-0.6cm}
\section{Proposed method}
\vspace{-0.2cm}
\subsection{Overall structure}
\vspace{-0.05cm}
In horizontal FL, the feature space is shared among all client datasets $\mathcal{D}_i=\{(\textbf{x}_k^i, y_k^i)\}_{k=1}^{m_i}$, $i=1,...,N$. The local model owned by each participant has the same structure $f$ as the global model. In a communication round $t$, first, each client optimizes its local model parameters $\bm{\theta}_i^t$ with $\mathcal{D}_i$ and sends the parameter gradients $\nabla\bm{\theta}_i^t$ (or $\bm{\theta}_i^t$, the two are equivalent in FL) to the central server. The central server then collects and aggregates $\bm{\theta}_i^t$'s to update the global model by
\begin{flalign}
    \label{eq:fed_sgd}
    \bm{\theta}_G^{t+1} \leftarrow \bm{\theta}_G^t -  \frac{\alpha_{\bm{\theta}}}{N}\sum_{i=1}^{N}\nabla\bm{\theta}_i^t,
\end{flalign}
where $\alpha_{\bm{\theta}}$ is the predefined learning rate.
Finally, the renewed model parameters $\bm{\theta}_G^{t+1}$ are broadcast to all participants.
This is the workflow of the basic FederatedAverage (FedAvg) algorithm.
In this process, the central server cannot access the raw data held by each client. Therefore, for clients, the value of data is reflected by the gradients they upload to central server. If the contribution of each gradient to the global model can be evaluated, it will indirectly reflect the client's contribution.

\begin{figure}[htp]
\begin{minipage}[b]{1.0\linewidth}
  \centering
  \centerline{\includegraphics[width=6.5cm]{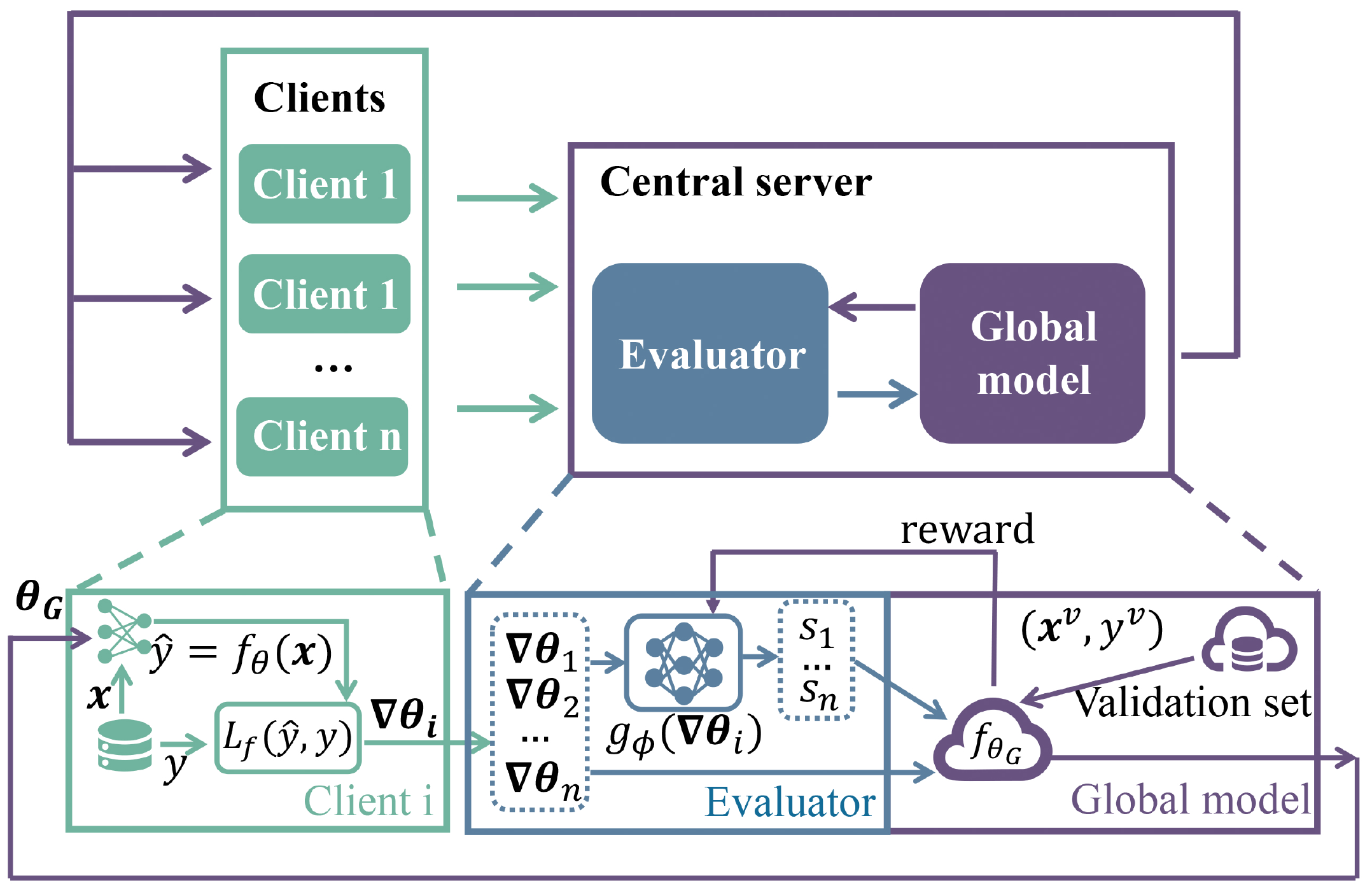}}
\end{minipage}
\caption{The block diagram of F-RCCE method.}
\label{fig1:framework}

\end{figure}

To achieve this goal, a novel method is proposed, named Federated REINFORCE client contribution evaluation (F-RCCE).
The F-RCCE is an integrated evaluation algorithm with the horizontal FL system (Fig.\ref{fig1:framework}).
In the proposed framework, an evaluator module $g_{\bm{\phi}}$ is implemented on the central server. 
Before aggregating client updates, the central server runs the evaluator to obtain an estimate of the values of the gradients.
The gradients are then selected or discarded as decided by the evaluator.
Only the selected client gradients are aggregated to renew the global model.
In this framework, the evaluator is trained alongside the global task model $f_{\bm{\theta}}$, using a held-out validation set $\mathcal{D}_v=\{(\textbf{x}_k^v, y_k^v)\}_{k=1}^{m_v}$ owned by the central server.
Thus the evaluator is able to perform task-specific evaluation that serves designated data distribution.

Optimization of the evaluator defines a non-differentiable sequential decision-making task.
It is difficult to be achieved by end-to-end gradient descent.
This problem is modeled as a RL problem for decision-making \cite{sutton2018reinforcement,mnih2015human, silver2016mastering}.
Details on the proposed formulation is explained in Section \ref{sec:f-rcce}.

In this paper, we assume a modest security environment for the FL system.
That is, all participants of the system are honest-but-curious.
Client datasets $\mathcal{D}_i$ are private to the respective participants.
The validation set $\mathcal{D}_v$ is selected to the need of the task model owner and private to the central server.
All communications to and from the central server should be encrypted.
In the experiments the encryption / decryption is omitted because it is independent of the proposed methods.

\vspace{-0.4cm}
\subsection{F-RCCE} \label{sec:f-rcce}
In F-RCCE, the evaluator module is formulated as an RL agent that collects client updates $\bm{\theta}_i^t$, and outputs a selection probability $\bm{\omega}^t$, that is, $g_{\bm{\phi}}(\bm{\theta}^t)=\bm{\omega^t}$.
The evaluator then samples a selection vector $\textbf{S}^t=[s_1^t, ... ,s_n^t]$, where $s_i^t\in\{0, 1\}$ represents $\bm{\theta_i^t}$ being included or discarded in the model aggregation.
$P(s_i^t=1)=\bm{\omega}_i^t$ is the probability of $\bm{\theta_i^t}$ being included. 
After model aggregation, a reward signal $r$ is calculated based on the global model's performance on $\mathcal{D}_v$.
In summary, the proposed RL problem is:
\begin{itemize}
\item {\it State space}: The feasible set of $\bm{\theta}_G$.\vspace{-0.2cm}
\item {\it Action space}: $\textbf{S}^t$.\vspace{-0.2cm}
\item {\it Reward function}: $r(\textbf{S}^t)$.
\end{itemize}

In this paper, the reward function is defined to be directly related with the global model's performance on the validation set.
Specifically,
\vspace{-0.3cm}
\begin{flalign}
    \label{equ2}
    r(\textbf{S}^t) = \frac{1}{m_v}\sum_{k=1}^{m_v}L_v\left(f_{\bm{\theta}_G}(\textbf{x}_k^v), y_k^v\right)-\delta,
\end{flalign}
where $L_v$ is the loss function of global model on validation set $\mathcal{D}_v$, and $\delta$ is a baseline calculated by moving average of previous loss $L_v$ with moving averaging window $T>0$.

As opposed to FedAvg, the global model parameters are updated as
\vspace{-0.5cm}
\begin{flalign}
    \label{equ1}
    \bm{\theta}_G^{t+1} \leftarrow \bm{\theta}_G^t -  \frac{\alpha_{\bm{\theta}}}{\sum_{i=1}^{N}s_i^t}\sum_{i=1}^{N}s_i^t\nabla\bm{\theta}_i^t.
\end{flalign}

\vspace{-0.2cm}To optimize the evaluator, the objective function of $g_{\bm{\phi}}$ is defined as follows:
\begin{flalign}
J(\bm{\phi}) &= \mathbb{E}_{\textbf{S}^t\sim\pi_{\bm{\phi}}(\bm{\theta}^t,\cdot)}r(\textbf{S}^t),
\end{flalign}
where $\pi_{\bm{\phi}}(\bm{\theta}^t,\cdot)$ is a stochastic, parameterized policy defined by $\bm{\phi}$. According to the policy gradient theorem \cite{sutton2000policy}, we can obtain the probability of $\textbf{S}^t$ by $p(\textbf{S}^t|\bm{\phi})=\prod_{i=1}^n[{\bm{\omega}_i^t}^{s_i^t}\cdot (1-{\bm{\omega}_i^t})^{1-s_i^t}]$. With the log-derivative trick, we have
\begin{flalign}
    \nabla_{\bm{\phi}}p(\textbf{S}^t|\bm{\phi})=p(\textbf{S}^t|\bm{\phi})\nabla_{\bm{\phi}}\log p(\textbf{S}^t|\bm{\phi}),
\end{flalign}
where
\vspace{-0.3cm}
\begin{flalign}
    \nabla_{\bm{\phi}}\log p(\textbf{S}^t|\bm{\phi})=\sum_{i=1}^n s_i^t\nabla_{\bm{\phi}} \log \bm{\omega}^t_i+(1-s_i^t)\nabla_{\bm{\phi}}\log(1-\bm{\omega}^t_i).
\end{flalign}
Putting it all together, the following policy gradient can be given as
\begin{flalign}
    \label{equ:policygrad}
     \nabla_{\bm{\phi}}J(\bm{\phi})=\mathbb{E}_{\textbf{S}^t\sim\pi_{\bm{\phi}}(\bm{\theta}^t,\cdot)}\nabla_{\bm{\phi}}\log p(\textbf{S}^t|\bm{\phi})r(\textbf{S}^t).
\end{flalign}
Then, the evaluator's model parameters $\bm{\phi}$ can be optimized by gradient ascent method with learning rate $\alpha$:
\vspace{-0.2cm}
\begin{flalign}
    \label{equ:ascgrad}
    \bm{\phi}^{t+1} \leftarrow \bm{\phi}^{t} + \alpha_{\bm{\phi}}\sum_{i=1}^n r(\bm{S}^t)\nabla_{\bm{\phi}}\log p(\textbf{S}^t|\bm{\phi})|_{\bm{\phi}^t}.
\end{flalign}

The pseudo-code for the proposed F-RCCE algorithm is illustrated in  Algorithm \ref{algo:F-RCCE}.

\SetKwInput{KwIn}{Inputs}
\SetKwInput{KwInit}{Initialize}
\begin{algorithm}[htbp]
    \label{algo:F-RCCE}
  \SetAlgoLined
  
  \KwIn{Learning rate $\alpha_{\bm{\theta}}, \alpha_{\bm{\phi}} > 0$, moving average window $T > 0$}
  
  \KwInit{Parameters $\bm{\theta_G^0}, \bm{\phi^0}$, moving average $\delta$}
  \For{round $t=1,2,...$ }{
    \textbf{Clients execute:}{\\
            \For{client $c=1, 2,..., N$}{
                Client $c$ copies $\bm{\theta}^{t-1}_G$ as local model $\bm{\theta}^{t-1}_c$\;
                Client $c$ updates local model $\bm{\theta_c^t}$;
            }
        }
    \textbf{Server executes:}{\\
        Collect $\bm{\theta}^t=[\bm{\theta}_1^t, ... ,\bm{\theta}_N^t]$\;
        Calculate selection probabilities $\bm{\omega}^t=g_{\bm{\phi}}(\bm{\theta}^t)$\;
        Sample selection vector $\textbf{S}^t\sim Ber(\bm{\omega}^t)$\;
        Update $\bm{\theta_G^{t-1}}$ with Eq.(\ref{equ1})\;
        Calculate reward signal $r(\textbf{S}^t)$ with Eq.(\ref{equ2})\;
        Calculate gradient of $\bm{\phi}^t$ with Eq.(\ref{equ:policygrad})\;
        Update $\bm{\phi}^t$ with Eq.(\ref{equ:ascgrad})\;
        Update $\delta \leftarrow \frac{T-1}{T}\delta + \frac{1}{T} L_v(f_{\bm{\theta}_G}(\textbf{x}^v), y^v)$
        }
    }
  \caption{F-RCCE Algorithm.}
\end{algorithm}

\label{sec:pagestyle}
\vspace{-0.5cm}
\section{Experiments}
\label{sec:typestyle}
\vspace{-0.12cm}
\subsection{Experimental settings}
In this section, a Horizontal FL environment with a large number of clients ($\geq 50$) is simulated to test the proposed F-RCCE method on the SMS Spam dataset \cite{almeida2011contributions}. The SMS Spam data set contains 5572 samples, among which 5000 samples are divided into 50 groups according to Dirichlet distribution and distributed to each client, and the remaining 572 samples are stored as a validation set in the central server. The task model $f_{\bm{\theta}}$ classifies SMS as ham or spam. In data preprocessing, the data is tokenized and the text is converted to sequences using Keras with $max\_words=1000$ and $max\_len=150$. After that, data was standardized to have zero mean and one standard deviation using scikit-learn's Standard Nomalizer. Labels of categorical data are encoded in one-hot embedding.

On the model selection, logistic regression classifier is used as the task model and a four-layer multi-layer-perceptron as evaluator model. Adam optimizer with learning rate $\alpha_{\bm{\phi}}=10^{-5}$ is used to optimize evaluator model and SGD optimizer with learning rate $\alpha_{\bm{\theta}}=0.1$ is used to optimize the local model which each client copied by global model.

To verify the authenticity of the reported contribution factor, the remove-and-retrain scheme is adopted which proposed by Yoon et al \cite{yoon2019data}. 
Specifically, given an estimate of contribution values, the lowest- or highest-valued datapoints are discarded.
The task model are then re-trained with the remaining data.
When the value estimation is correct, removing high-valued datapoints should lead to a decrease in the task model accuracy, vice versa.

\begin{figure*}[htpb]
\begin{minipage}[b]{0.24\linewidth}
  \centering
  \centerline{\includegraphics[width=3.8cm]{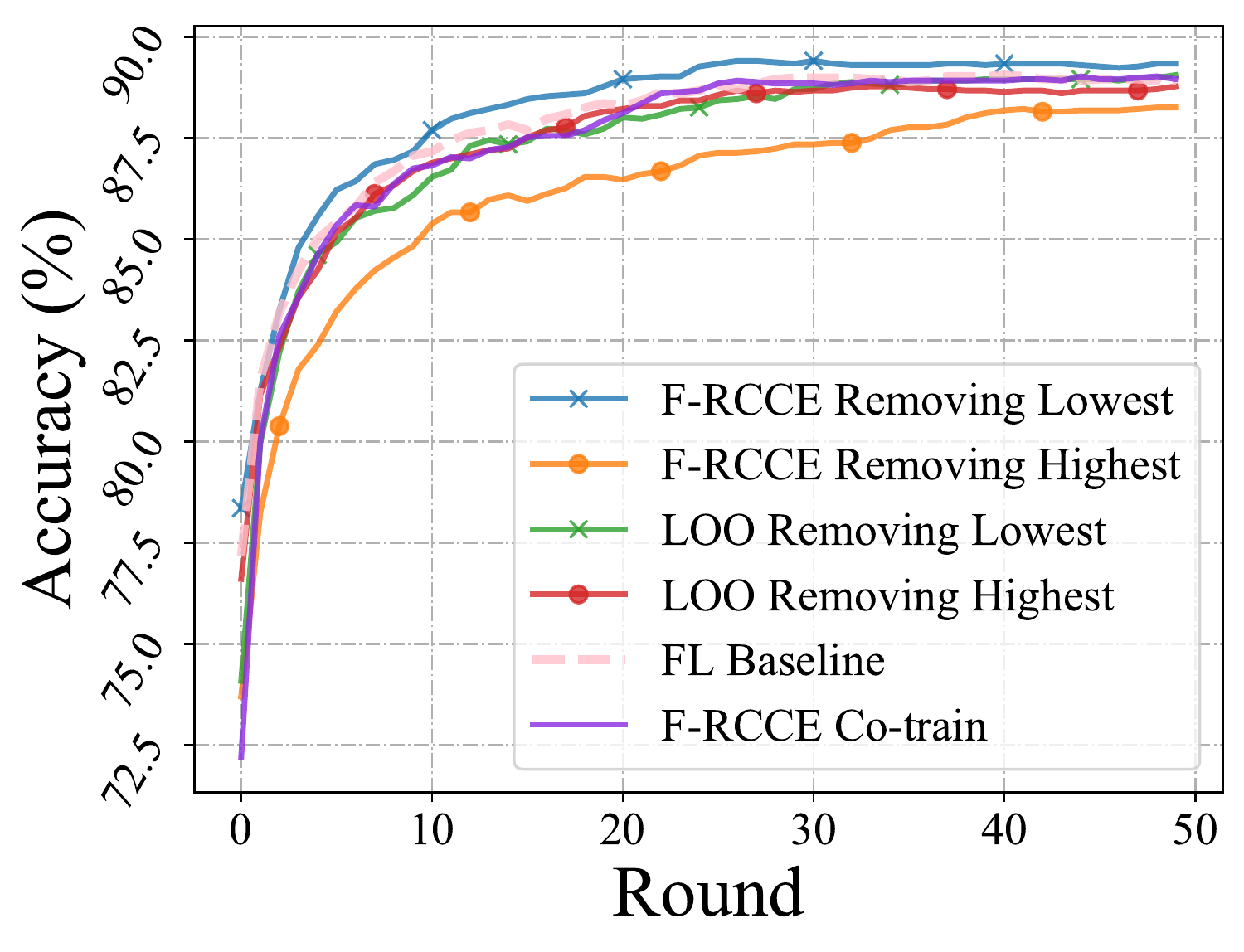}}
  \centerline{(a)}\medskip
\end{minipage}
\begin{minipage}[b]{0.24\linewidth}
  \centering
  \centerline{\includegraphics[width=3.81cm]{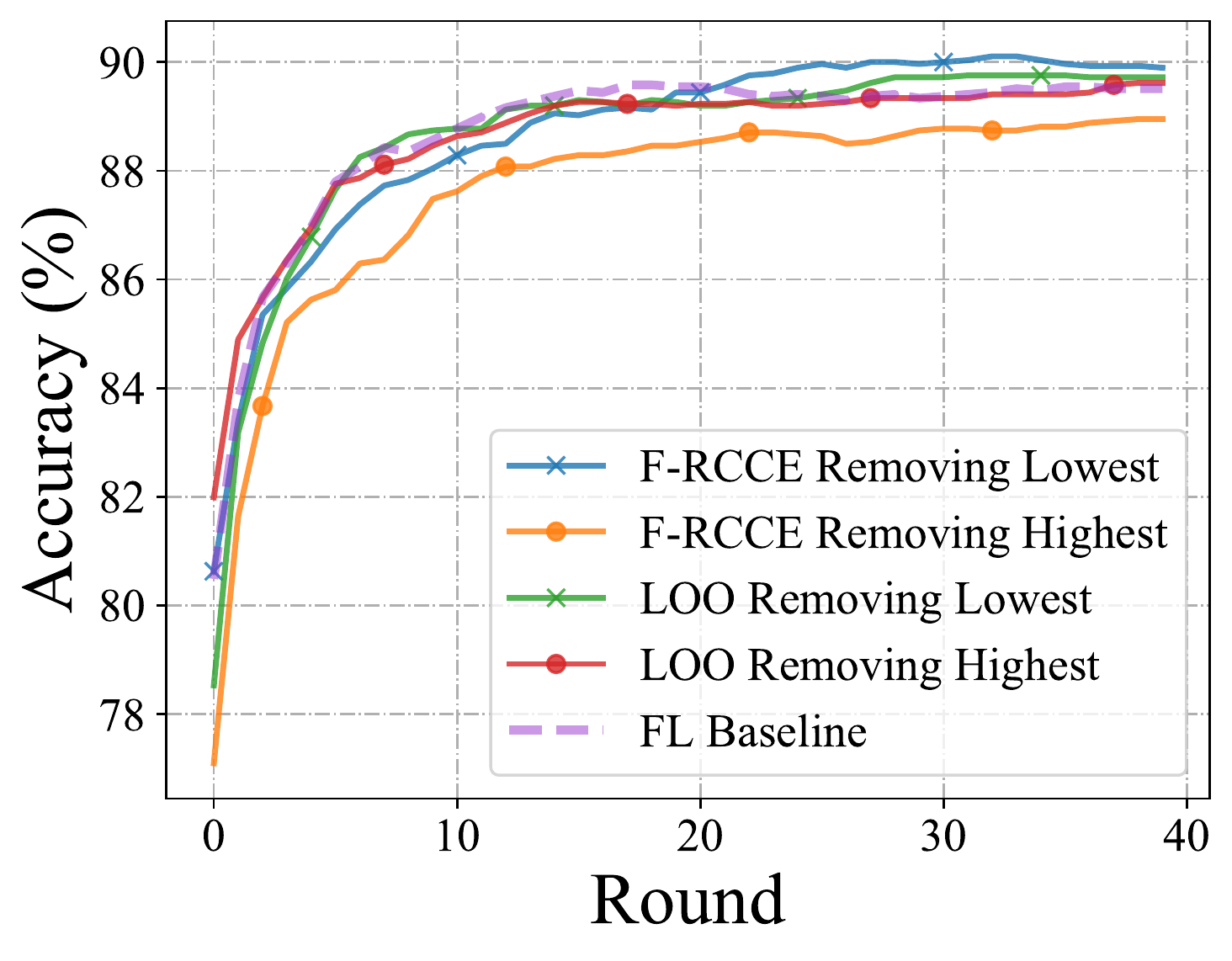}}
  \vspace{-0.01cm}
  \centerline{(b)}\medskip
\end{minipage}
\begin{minipage}[b]{0.24\linewidth}
  \centering
  \centerline{\includegraphics[width=3.81cm]{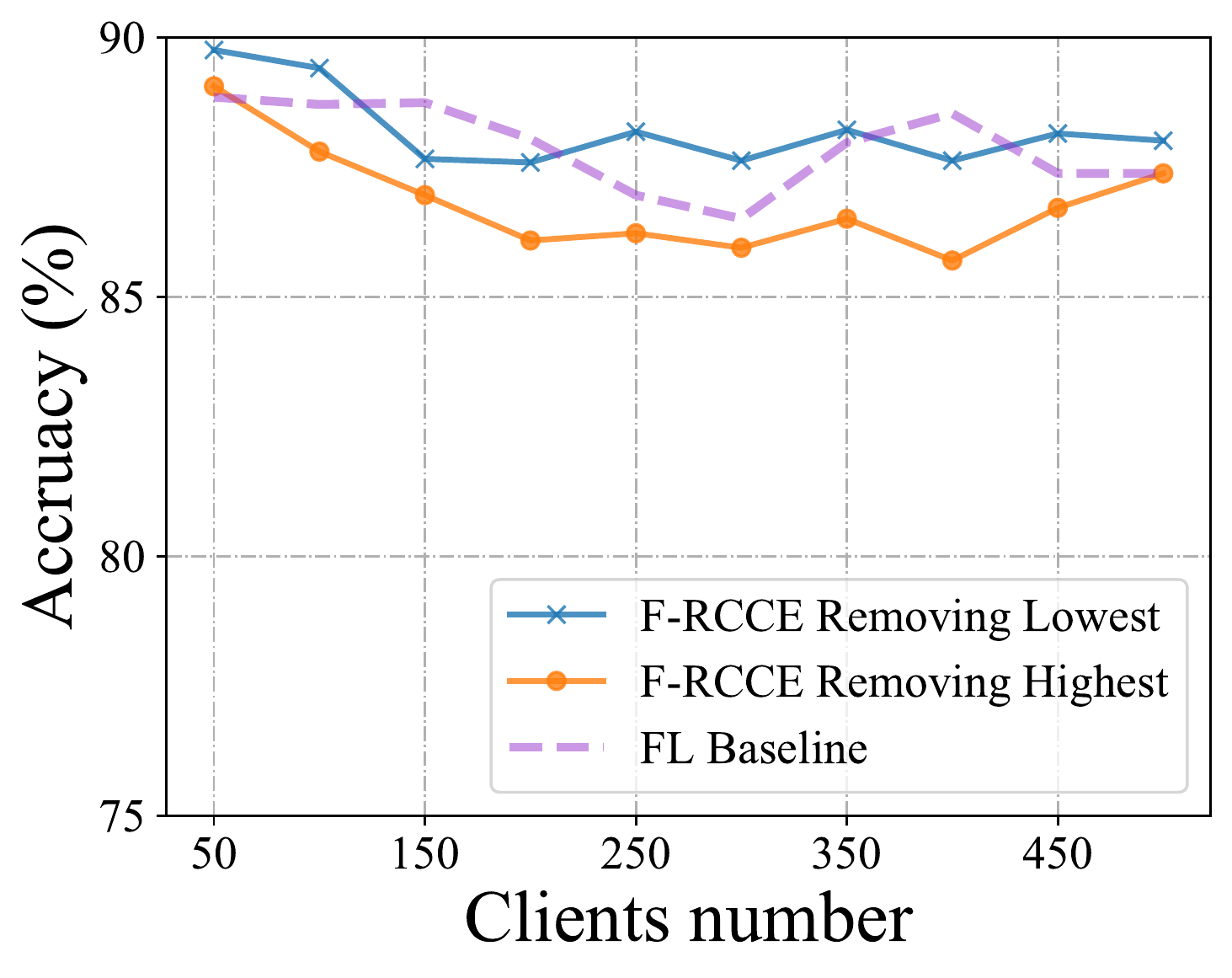}}\vspace{-0.04cm}
  \centerline{(c)}\medskip
\end{minipage}
\begin{minipage}[b]{0.24\linewidth}
  \centering
  \vspace{0.05cm}
  \centerline{\includegraphics[width=3.425cm]{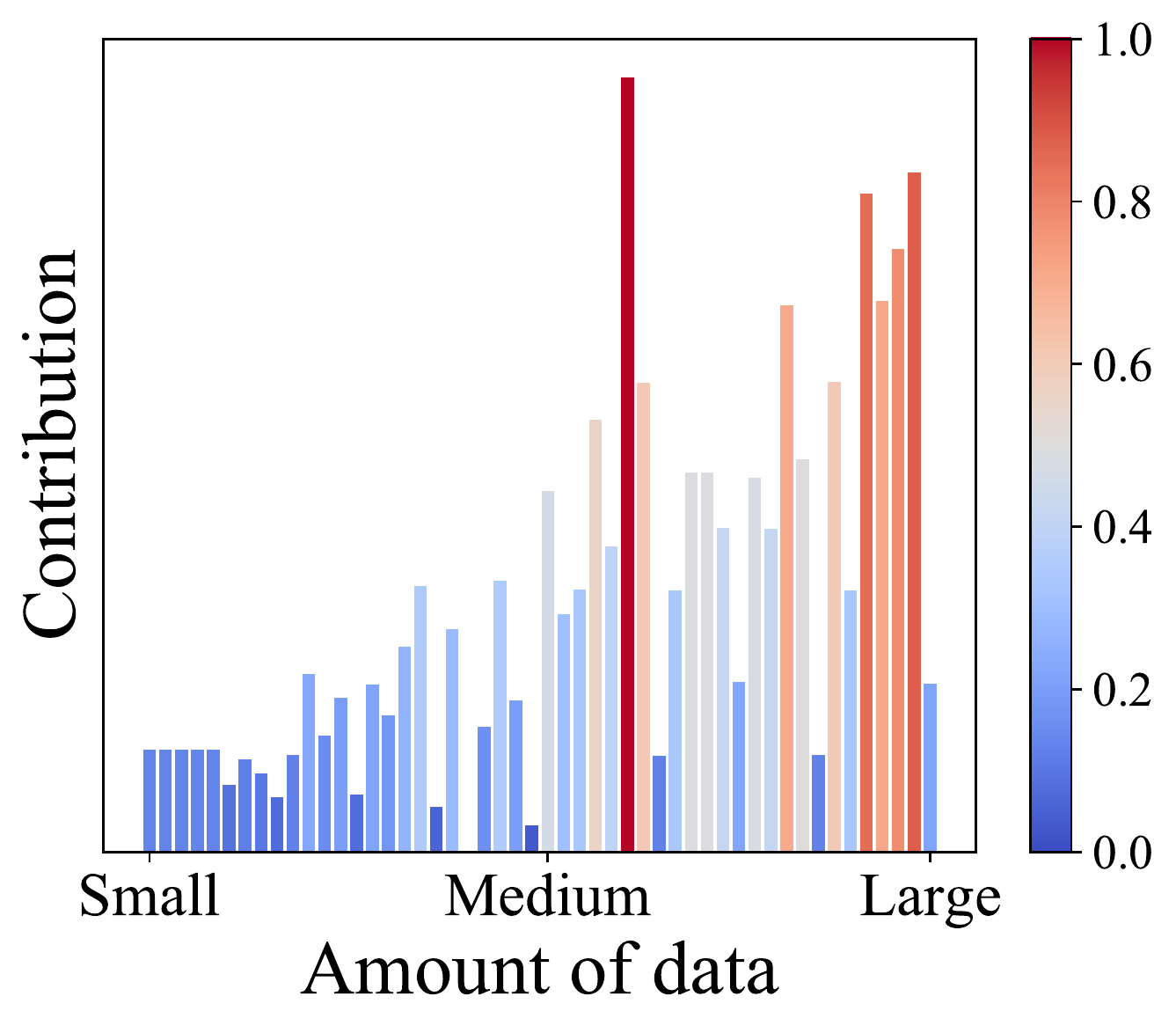}}
  \centerline{(d)}\medskip
\end{minipage}
\vspace{-0.5cm}\caption{Experimental results. (a) shows the result of F-RCCE and LOO of removing highest-/lowest-contribution gradient; (b) shows the performance of F-RCCE and LOO of removing highest-/lowest-contribution gradient with noisy data; (c) shows the performance of F-RCCE with different number of clients; (d) shows the contribution of clients measured by F-RCCE.}
\label{fig2:exp}
\end{figure*}

In order to reduce the uncertainty caused by random initialization, each experiment is repeated five times. The experimental data is averaged as the presented results.
The baseline task model is trained by conventional FedAvg, without any data value estimation. 
The convergence curves of the FedAvg baseline as well as the co-trained F-RCCE model are presented in Fig.\ref{fig2:exp}(a).
The two curves are almost identical, demonstrating that the inclusion of the evaluator $\bm{\phi}$ does not have a noticeable impact on the federated model.

\vspace{-0.25cm}
\subsection{Comparison with LOO}
\vspace{-0.15cm}
As aforementioned, LOO\cite{wang2019measure} cannot directly measure the value of gradients uploaded by clients for each communication round, a specific client is deleted in each re-training iteration to evaluate the contribution of individual client. Afterwards, part of the highest-/lowest-contribution clients are removed following removing rate in re-training stage. 

For F-RCCE, the evaluator and the global task model are trained for 1,000 communication rounds. 
Then the task model is re-trained with the evaluator fixed. 
In the re-training process, the selection probability output by the evaluator is taken as the contribution value of the client update.
At each round, 30\% of the client gradients are removed with the highest (F-RCCE Removing Highest) or lowest (F-RCCE Removing Lowest) contribution values.
The task model is updated with the remaining gradients.
The validation accuracy of the task model is recorded for each round. 

As shown in Fig.\ref{fig2:exp} (a), removing the highest contribution gradients makes the model converge more slowly than the baseline. 
Removing the lowest contribution gradients, on the other hand, slightly improves the validation accuracy.
In comparison, the difference between removing highest or lowest valued clients by the LOO method is not significant.

\vspace{-0.15cm}
\subsection{Experiments with corrupted data}
\vspace{-0.1cm}
To verify the ability of F-RCCE to identify data abnormality in horizontal FL, a training set is generated with randomly corrupted data. 
Noises is added to 20\% of the SMS samples by deleting 20\% of the words in each text string.
The subsequent data preprocessing is same as the previous experiment. 

Experimental results are shown as Fig.\ref{fig2:exp} (b). 
The LOO method again fail to distinguish the highest-/lowest-contribution clients. 
The F-RCCE method performs significantly better than LOO. It is more sensitive to the removal of high/low contribution gradients. After removing the gradients with highest contribution calculated by F-RCCE, the performance of the global model has an obvious decline. And the performance of the global model is gratifying after removing the gradients with lowest contribution calculated by F-RCCE.

\vspace{-0.4cm}
\subsection{Impact of the number of clients}
\vspace{-0.15cm}
In a business-to-client model, horizontal FL often has a large number of clients.
The following experiment investigates the consistency and time complexity of F-RCCE verses increasing number of clients.
Horizontal FL systems are simulated with 50 to 500 clients. Considering the convergence of the Evaluator requires a number of iterations, each global model in F-RCCE iterates for 1000 rounds. As a control, each global model iterates for 50 rounds under LOO. The running time of the two algorithms are recorded, as shown in Table \ref{tab:times}. 
With the number of clients increasing from 100 to 500, the time cost of F-RCCE only increases by 6\%. However, the time cost of LOO increased almost linearly. Therefore, F-RCCE is more applicable in in practical situations.

The validation accuracies with highest-/lowest-contribution gradients verses different numbers of clients are plotted on Fig.\ref{fig2:exp} (c).
It shows that, removing the highest-contribution gradients always degrades task model performance. Removing the lowest-contribution gradients can sometimes benefit the validation accuracy, but sometimes causes destructive consequences. This phenomenon is due to the fact that the datasets are randomly generated. Some of the removed gradients still have positive effects on the model, although its contribution ranks low compared to others.
There is no obvious trend in the validation accuracy with the increased number of clients. Result support that F-RCCE performs consistently in both small and large numbers of clients.
\begin{table}[htpb]
    \centering
    \begin{tabular}{cccccc}
        \toprule
            Clients number & 100 & 200 & 300 & 400 & 500 \\
         \hline
            LOO (s) & 249.5 & 464.6 & 699.6 & 806.1 & 1135.9\\
            F-RCCE (s) &62.6 & 63.0 & 64.2 & 64.9 & 66.9\\
         \bottomrule
    \end{tabular}
    \caption{Time cost of LOO and F-RCCE on different number of clients.}
    \label{tab:times}
\end{table}

\vspace{-0.4cm}
\subsection{Client contribution using F-RCCE}
\vspace{-0.2cm}
A intuitive idea is proposed to measure the clients contribution using F-RCCE. That is, the selection probability calculated by F-RCCE in each iteration is taken as the contribution of the iteration, and the contributions in all iterations are summed and scaled as the client contribution. For clients, when their amount of clean data is large, the calculated gradient is relatively stable, and it is easier to eliminate the influence of outlier. So the amount of clean data is considered to be positively correlated with the client's contribution. In the experiment, a trained evaluator is used to measure the clients contribution for 50 communication rounds. Clients are sorted by the number of samples they held, and the contributions are plot in Fig. \ref{fig2:exp} (d). With the increase of the number of samples, the contribution of client also showed an upward trend. Experimental result shows that F-RCCE has the ability to evaluate the contribution of clients.

\vspace{-0.4cm}
\section{CONCLUSION}
\label{sec:refs}
\vspace{-0.3cm}
Fair and accurate measurement of the contribution of each parcitipant is important for the FL society. 
In this work, a client contribution evaluation method named F-RCCE is proposed which accomplished integrated contribution evaluation using RL method.
Experimental results strongly support that the proposed method can accurately evaluate the contribution of gradients provided by each client. 
Its time cost remained almost constant with increased number of clients, so it is suitable for applications with large business-to-client models.
Future research may consider how to effectively distribute benefits to all participants using the clients' contribution obtained by F-RCCE. This is one of the key components in attempts to instantiate the incentive design for the FL ecosystem.

\vspace{-0.5cm}
\section{ACKNOWLEDGEMENTS}
\vspace{-0.3cm}
This paper is supported by National Key Research and Development Program of China under grant No. 2018YFB1003500, No. 2018YFB0204400 and No. 2017YFB1401202. Corresponding author is Jianzong Wang from Ping An Technology (Shenzhen) Co., Ltd.
\bibliographystyle{IEEEbib}
\bibliography{strings,refs}

\end{document}